\documentclass[runningheads]{llncs}

\usepackage{eccv}

\usepackage{eccvabbrv}

\usepackage{graphicx}
\usepackage{booktabs}
\usepackage{algorithm,algorithmic}
\usepackage{multirow}
\usepackage{makecell}

\usepackage[accsupp]{axessibility}  

\usepackage{hyperref}

\usepackage{orcidlink}

\begin{document}

\title{Enhancing Tracking Robustness with Auxiliary Adversarial Defense Networks} 

\titlerunning{Enhancing Tracking Robustness with AADN}

\author{Zhewei Wu\orcidlink{0000-0003-1488-8728} \and
Ruilong Yu\and
Qihe Liu$^\dag$\orcidlink{0000-0002-8195-1304}\and
Shuying Cheng\orcidlink{0009-0002-3275-589X}\and
Shilin Qiu\and
Shijie Zhou}

\renewcommand{\thefootnote}{}
\footnotetext[2]{$^\dag$ Corresponding author.}

\authorrunning{Z.~Wu, R.~Yu, et al.}

\institute{University of Electronic Science and Technology of China, Chengdu, 610000, China \\
\email{\{wuzhewei0914,shuyingcheng0214\}@foxmail.com, yrl666@outlook.com,
	\{qiheliu,sjzhou\}@uestc.edu.cn, qiushilin@std.uestc.edu.cn}}

\maketitle

\begin{abstract}
Adversarial attacks in visual object tracking have significantly degraded the performance of advanced trackers by introducing imperceptible perturbations into images. However, there is still a lack of research on designing adversarial defense methods for object tracking. To address these issues, we propose an effective auxiliary pre-processing defense network, AADN, which performs defensive transformations on the input images before feeding them into the tracker.  Moreover, it can be seamlessly integrated with other visual trackers as a plug-and-play module without parameter adjustments. We train AADN using adversarial training, specifically employing Dua-Loss to generate adversarial samples that simultaneously attack the classification and regression branches of the tracker. Extensive experiments conducted on the OTB100, LaSOT, and VOT2018 benchmarks demonstrate that AADN maintains excellent defense robustness against adversarial attack methods in both adaptive and non-adaptive attack scenarios. Moreover, when transferring the defense network to heterogeneous trackers, it exhibits reliable transferability. Finally, AADN achieves a processing time of up to 5ms/frame, allowing seamless integration with existing high-speed trackers without introducing significant computational overhead. 
\keywords{Adversarial Defense \and Object Tracking \and Adversarial Training}
\end{abstract}

\section{Introduction}
\label{sec:intro}

Visual object tracking predicts the location of a target in a continuous sequence of video frames based on the specified appearance of target in the initial frame. As a fundamental task in computer vision, it has received extensive attention and finds wide-ranging applications in various domains such as daily life and industry~\cite{autodrive1,autodrive2,robot1,robot2}. Research on adversarial samples~\cite{szegedy2014} has revealed the ability to deceive deep neural networks by introducing imperceptible perturbations into images. Consequently, an increasing number of researchers have turned their focus towards adversarial attack methods specifically for object tracking~\cite{spark,csa,iouattack,uta,eusa}. These methods employ elaborate objective functions to guide network models~\cite{csa,eusa,dfa} or iterative approaches~\cite{spark,iouattack} to generate imperceptible adversarial perturbations that cause trackers to lose track of the target. Undoubtedly, enhancing the adversarial robustness of DNN-based trackers is an urgent problem.

Unfortunately, there is still a lack of research and significant challenges in designing adversarial defense methods for object tracking.
Firstly, most existing defense methods are specifically designed for image classification~\cite{freeat,fastat,defdistillation}. Due to the structural differences between classifiers (typically backbone + FClayers) and trackers (typically backbone + matching + prediction), and SPARK~\cite{spark} has demonstrated difficulty of transfering attack methods between them. Secondly, adversarial training, as an efficient defense method, is commonly used to enhance model robustness~\cite{atreview}. However, it requires retraining the entire network when apply defense mechanism to tracker, which imposes significant time and computational costs. Lastly, the continuity of tracking tasks implies that defense methods working in conjunction with trackers should not significantly degrade the tracking speed. In other words, the performance improvement brought  by the defense method should not outweigh the computational overhead it introduces.

To address these challenges, we propose a novel defense network called AADN for object tracking. The network architecture is based on the U-Net~\cite{unet}, and the network parameters are optimized by Dua-Loss guided adversarial training. Specifically, Dua-Loss consists of fore-background classification loss that reinforces coarse target localization and regression loss that refines the precise target position. The adversarial training approach allows thorough exploration of potential adversarial samples within the clean sample neighborhood based on gradient information, thus constructing a defense network with stronger generalization capabilities. Importantly, AADN is designed as an auxiliary module placed before the siamese backbone and can be seamlessly integrated with various advanced trackers in a plug-and-play manner. This enables model deployers to avoid the significant time consumption of retraining the whole model and allows targeted defense of the template and search regions based on specific needs.

\begin{figure}[h]
	\centering
	\includegraphics[width= 0.7\linewidth]{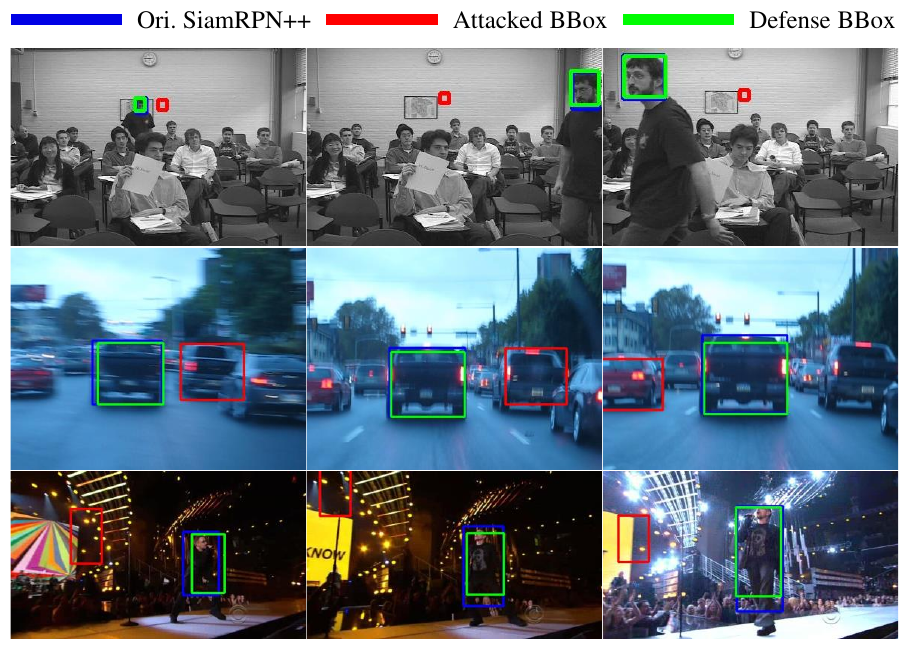}
	\caption{Visualized results of defense performance on 3 sequences from OTB100. Better viewed in color with zoom-in.}
	\label{vis-bbox}
\end{figure}

Extensive experiments on the OTB100~\cite{otb2015}, VOT2018~\cite{vot2018}, and LaSOT~\cite{lasot} benchmarks demonstrate the strong robustness against adversarial attack methods in both adaptive and non-adaptive attack scenarios. It also exhibits excellent defense transferability across different heterogeneous trackers, even without retraining.\cref{vis-bbox} illustrates that AADN maintains stable defense generalization in challenging scenarios, including similar object, background clutter, deformation, and illumination variance. In terms of defense efficiency, our defense network achieves the processing speed of 5ms/frame, allowing it to be used in conjunction with existing high-speed trackers without significantly reducing the tracking speed.

In summary, our contributions can be highlighted in three aspects:

1. To the best of our knowledge, we are the first to conduct research specifically on adversarial defense for object tracking. The proposed AADN demonstrates excellent defense robustness against adversarial attack methods in both adaptive and non-adaptive attack scenarios.

2. AADN exhibits outstanding defense transferability, enabling seamless integration with existing trackers without retraining. This enhances the robustness of the trackers against adversarial attacks in a plug-and-play manner.

3. AADN demonstrates excellent processing efficiency, with a speed of 5ms /frame for defense sample processing. When used in conjunction with existing high-speed trackers, AADN does not introduce significant computational overhead.

\section{Related Works}

\subsection{Adversarial Attacks in Object Tracking}

Since the discovery of adversarial samples~\cite{szegedy2014}, researchers have conducted extensive studies on adversarial sample generation~\cite{fgsm,csa} and defense~\cite{freeat,fastat,defdistillation,gramask}. In recent years, some researchers have also explored adversarial sample generation methods specifically for object tracking tasks. 

SPARK~\cite{spark} demonstrates that classical BIM~\cite{bim}, and C\&W~\cite{cw} methods have poor transferability to tracking task and proposes an online incremental attack method that utilizes information from past frames. One-Shot Attack~\cite{oneshot} attacks the initial frame before the tracking process begins using batch confidence loss and feature loss. FAN~\cite{fan} designs an efficient and fast attack method that induces drift loss and embedded feature loss to deviate the predicted bounding box from the true position and guide it towards a specific trajectory. CSA~\cite{csa} suppresses the target region in the heat map using a cooling-shirinking loss and induces the tracker to contract the bounding box. IoU Attack~\cite{iouattack} obtains the adversarial sample with the lowest IoU score by employing an orthogonal approximation method in a black-box iterative manner with minimal noise. DFA~\cite{dfa} generates adversarial perturbation by minimizing the difference between foreground and background regions in feature space. UTA~\cite{uta} devises a universal adversarial perturbation that can be directly transferred to other tracking networks and datasets to achieve targeted attacks. It is evident that the majority of trackers experience significant degradation in tracking accuracy when confronted with adversarial samples. Therefore, conducting research to ensure that trackers remain robust against adversarial attacks is of paramount importance.

However, to the best of our knowledge, there is currently no research on adversarial defense specifically for object tracking. In this paper, we propose a adversarial defense network that is applicable to object tracking.

\subsection{Adversarial Training}

Adversarial training has been proven to be the effective method for mitigating adversarial attacks~\cite{atreview} and thus has received widespread attention in the research community. The intention of adversarial training is to continuously generate adversarial samples in each training batch, thereby augmenting the training data. Consequently, models that have undergone adversarial training exhibit greater robustness against adversarial samples compared to models trained with standard methods.

PGD-AT~\cite{pgdat} formulates the basic idea of adversarial training through a min-max optimization problem. The inner maximization problem aims to find the worst-case adversarial samples for the trained model, while the outer minimization problem aims to train a model that is robust to adversarial samples. Nevertheless, the limitation of such adversarial training methods is that they require several iterations to generate adversarial samples, which lead to more time consumption. Therefore, Free-AT~\cite{freeat} leverages the gradient information during backpropagation to synchronously update model parameters and adversarial perturbations. Building upon this, Fast-AT~\cite{fastat} employs a random initialization FGSM strategy to generate adversarial perturbations, which avoids the substantial time overhead introduced by repeated PGD~\cite{pgdat} iterations while achieving effective training results.

However, these approaches require retraining the original model to enhance its robustness against adversarial samples. In the case of object tracking models, the network architecture is more complex compared to classification models, making it more challenging to retrain them and imposing higher hardware requirements. To address this issue, inspired by Fast-AT, we adopt a similar training methodology for the proposed AADN network. AADN is deployed as an additional module before the tracker and does not require parameter adjustments during the training phase of the tracking model.

\section{Proposed Method}
\begin{figure}
	\centering
	\includegraphics[width= \linewidth]{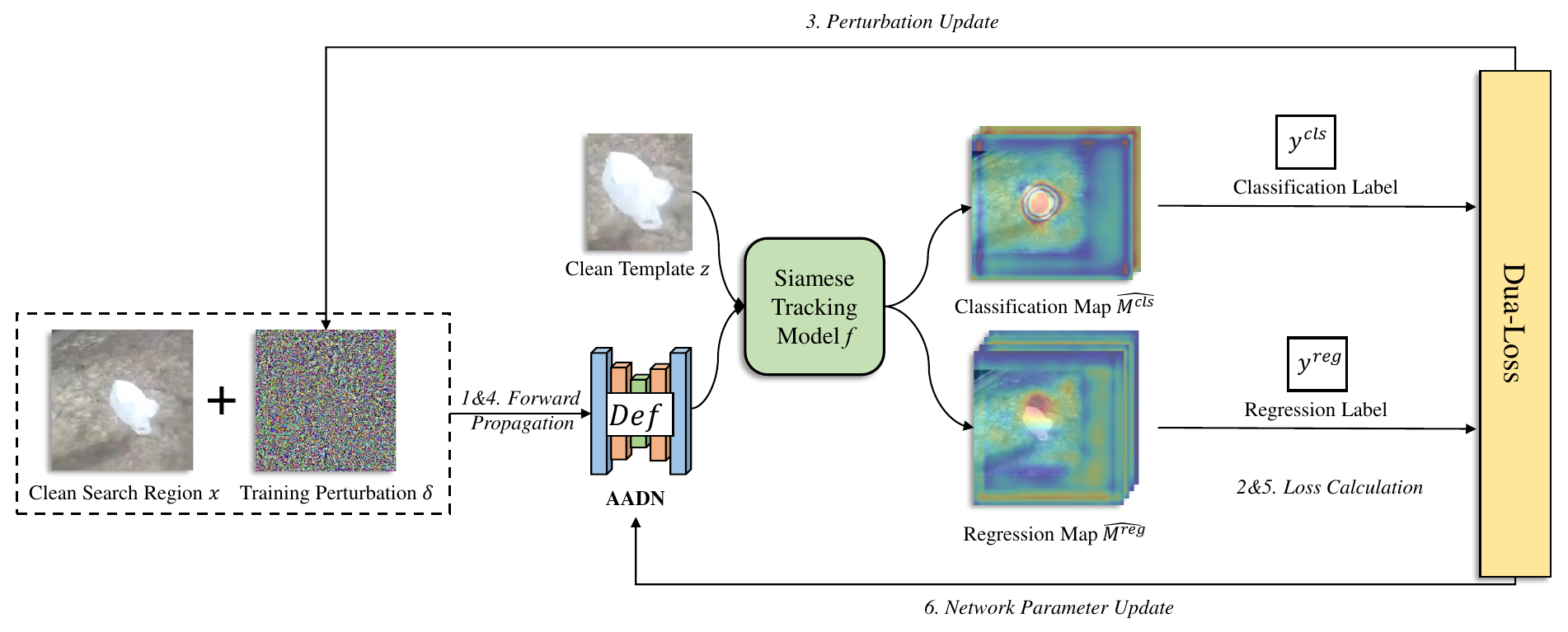}
	\caption{The training pipeline for deploying the proposed AADN on the search branch.}
	\label{pipeline}
\end{figure}

In this section, we present the proposed AADN network. Input images containing unknown perturbations are first input to the defense network for defensive transformation, then the transformed samples are passed to the tracking model for regular tracking. AADN consists of two independently trained components, which can be separately deployed in front of the search branch and the template branch of the tracker. As the operations in these two branches are highly similar, for the sake of brevity, the following explanation focuses on the search branch. The training pipeline for the proposed AADN network is illustrated in the \cref{pipeline}.

\subsection{Problem Definition}

Let ${ \left\{ I_i \right\}}_{i=1}^n$ represents an input image sequence, consisting of $n$ consecutive image frames. Additionally, let $\left\{ {b_i^{gt}} \right\} _{i=1}^n$ denotes the groundtruth bounding box in the $i$-$th$ frame. Here, the variable $b \in R^4 $ specifies the location and size of the target bounding box. Let $\left\{ y_i^{cls}, y_i^{reg}\right\}$ represents the classification and regression labels of $b_i^{gt}$ in $I_i$.

Consequently, the tracking process of siamese tracker could be summarized as follows: The pre-processing stage starts by obtaining the tracking template $z$ according to the initial frame $I_1$ and the annotation $b_1$. Subsequently, the search region $x_i$ is cropped from the consecutive frames $\left\{ I_i\right\}_{i=2}^n$ based on the previous tracking results $ \left\{ b^{pre}_{i-1}\right\}_{i=2}^n$. Then, the sample pair $\left\{ z, x_i \right\}$ is input into the tracking model $f(\cdot )$, resulting in the classification map $M_i^{cls}$ and the regression map $M_i^{reg}$, which contain information about the target's position, as defined in \cref{eq1}. During the post-processing stage, the tracker calculates the precise target bounding box $b_i^{pre}$  in frame $I_i$ based on $\left\{ M_i^{cls}, M_i^{reg} \right\}$.
\begin{equation}
	\left\{ M_{i}^{cls},M_{i}^{reg} \right\}=f\left ( z,x_{i} \right )
	\label{eq1}
\end{equation}
Based on the aforementioned definitions, the workflow of the AADN can be understood as follows: Firstly, the search region with unknown perturbations $x_i^{atk}$ are input into the defense network $Def(\cdot,\theta)$ with parameters $\theta$, resulting in the defense sample $\widehat{x_i} $. Together the clean template $z$, $\left\{ z, \widehat{x_i} \right\}$ is then fed into $f(\cdot )$ to obtain the defense classification map $\widehat{M_i^{cls}}$ and the defense regression map $\widehat{M_i^{reg}}$. 

During the adversarial training phase, our objective is to optimize $\theta$ by minimizing the Dua-Loss $L_{Dua}$ between $\left\{ \widehat{M_i^{cls}},\widehat{M_i^{reg}} \right\}$ and $\left\{ y_i^{cls}, y_i^{reg} \right\}$. This optimization objective can be formalized as presented in \cref{eq2}.
\begin{equation}
	\begin{aligned}
		&\underset{\theta}{min}L_{Dua}\left ( f\left ( z,Def(x_i^{atk},\theta) \right ),\left\{ y_i^{cls}, y_i^{reg} \right\} \right ) \\
		&= \underset{\theta}{min}L_{Dua}\left ( f\left ( z,\widehat{x_i} \right ),\left\{ y_i^{cls}, y_i^{reg} \right\} \right ) \\
		&= \underset{\theta}{min}L_{Dua}\left ( \left\{ \widehat{M_i^{cls}},\widehat{M_i^{reg}} \right\},\left\{ y_i^{cls}, y_i^{reg} \right\} \right )
	\end{aligned}
	\label{eq2}
\end{equation}

\subsection{Dua-Loss}
\label{sec-dualoss}

Dua-Loss consists of a fore-background classification loss $L_{cls}$, which enhances target coarse localization, and a regression loss $L_{reg}$ that refines the precise target position. In the proposed AADN, Dua-Loss serves two purposes: Firstly, it is used to evaluate the degree to which the tracker's output aligns with the groundtruth based on defense samples. Secondly, Dua-Loss is employed in the adversarial training process to generate adversarial samples simultaneously for both classification and regression tasks in each training batch.

As the tracking result $b^{pre}_i$ is directly computed from $\left\{ \widehat{M_i^{cls}},\widehat{M_i^{reg}} \right\}$ in the post-processing stage, The quality of $\left\{ \widehat{M_i^{cls}},\widehat{M_i^{reg}} \right\}$ directly determines the tracking accuracy. Moreover, since both classification and regression tasks are involved in tracking model, it is also crucial to generate adversarial samples that can attack both branches using the gradient information of Dua-Loss. This enables the AADN defense network to learn to filter adversarial perturbations from both branches. Dua-Loss follows the widely adopted standard tracking loss~\cite{siamrpn}, which can be formulated as \cref{eq3}.

\begin{equation}
	L_{Dua} = L_{cls} + L_{reg}
	\label{eq3}
\end{equation}

Here, $L_{cls}$ represents the cross-entropy loss. $L_{reg}$ is composed of the $SmoothL_1$ loss\cite{fastrcnn}. It can be expressed as \cref{eq4}.

\begin{equation}
	L_{reg} =  \sum_{t}^{\left\{ x,y,w,h \right\}} SmoothL_1(d_t,\sigma )
	\label{eq4}
\end{equation}

%
%
%
%
%
%

\subsection{Dua-Loss Guided Adversarial Training}

To address the widely adopted classification and regression branch structure in visual trackers, we employ Dua-Loss to generate adversarial samples with cross-task attack capability during the adversarial training process. Therefore, by combining \cref{eq2} and \cref{eq3}, we can define the following min-max problem to describe the objective of our adversarial training in \cref{eq9}:

\begin{equation}
	\underset{\theta}{min}\underset{{\left\|\delta  \right\|}_{2 }< \epsilon }{max}L_{Dua}\left ( \left\{ \widehat{M^{cls}},\widehat{M^{reg}} \right\},\left\{ y^{cls}, y^{reg} \right\} \right )
	\label{eq9}
\end{equation}

Here, $\delta$ represents the perturbation added to the clean search region $x$. The condition ${\left\|\delta  \right\|}_{2 }< \epsilon$ denotes the $l_2$-ball with center as the clean search region $x$ and radius as the perturbation budget $\epsilon$. The inner maximization problem involves finding the adversarial samples that lead to the worst-case prediction, while the outer minimization problem focuses on training network parameters that remain robust to adversarial samples. 

For each training batch, two forward passes are conducted in the network. In the first forward pass, AADN takes the search region with gaussian noise $x+\delta^g$ as input and outputs the defense search region $\widehat{x}$. After that, it is fed into the siamese tracking model $f$ along with the clean template $z$, which finally outputs the defense classification map $\widehat{M^{cls}}$ and the regression map $\widehat{M^{reg}}$. $\left\{ \widehat{M^{cls}},\widehat{M^{reg}} \right\}$ are used to calculate the Dua-Loss with the groundtruth $\left\{ y^{cls}, y^{reg} \right\}$ according to \cref{eq3}. At this point, we can update $\delta ^{g}$ to adversarial perturbation $\delta ^{adv}$ based on Dua-Loss according to \cref{eq10}:

\begin{equation}
	\delta ^{adv}= \delta ^{g}+\epsilon \cdot sign\left ( \bigtriangledown _{x} L_{Dua} \right )
	\label{eq10}
\end{equation}

In the second forward pass, the search region  with adversarial perturbation $x+\delta^{adv}$ and the clean template $z$ are used as inputs, and the Dua-Loss is recalculated in the same manner. Finally, the parameters $\theta$ of the defense network are updated through backpropagation. 

\begin{figure}[h!]
	\centering
	\includegraphics[width= 0.7\linewidth]{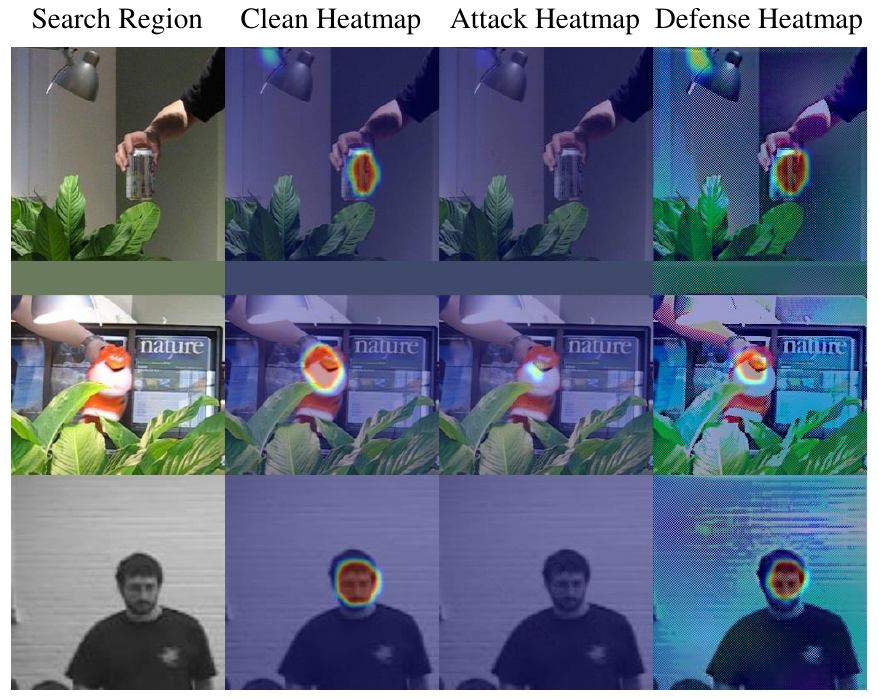}
	\caption{Visualized search regions from OTB100~\cite{otb2015} dataset and their corresponding heatmaps. The columns represent the types of visualized images, including the original search region, the heatmap after being attacked by the CSA method~\cite{csa}, and the heatmap obtained based on defense samples.}
	\label{vis-heatmap}
\end{figure}

In summary, the detailed adversarial training procedure is presented in \cref{alg1}. \cref{vis-heatmap} presents the visualized search regions and their corresponding heatmaps in both attack and defense scenarios. By examining the results in column \textit{Clean Heatmap}, \textit{Attack Heatmap} and \textit{Defense Heatmap}, it can be observed that the defense samples effectively restore the target's correct response in the search region. This enables the tracker to make accurate predictions.

\begin{algorithm}[h]
	\caption{Framework of adversarial training process of proposed AADN on search branch}
	\label{alg1}
	\begin{algorithmic}[1]
		\REQUIRE
		training dataset $D$, training epochs $T$, batch size $S$, perturbation budget $\epsilon$
		
		\ENSURE
		trained defense network parameters $\theta$
		
		\FOR{$i$ in $range[1, T]$}
		\FOR{random training batch $\left\{ x,z,y^{cls},y^{reg}\right\}\in D$ }
		
		\STATE Initialize training perturbation $\delta ^g$ with the gaussian noise. \\ $\delta^g \sim U\left [ -\epsilon ,\epsilon  \right ]$
		\STATE Obtain the defense sample $\widehat{x}$ by feeding $x+ \delta ^g$ into the defense network parameterized by $\theta$.\\ $\widehat{x}=Def(x+\delta ^g,\theta)$
		\STATE After the first forward pass, calculate the Dua-Loss with $\widehat{x}$.\\ $L_{Dua}\left ( f\left ( z,\widehat{x} \right ),\left\{ y^{cls}, y^{reg} \right\} \right )$
		\STATE Update $\delta ^g$ to adversarial perturbation $\delta ^{adv}$. \\ $ \delta ^{adv}= \delta ^{g}+\epsilon \cdot sign\left ( \bigtriangledown _{x} L_{Dua} \right )$
		\STATE Reobtain the defense sample $\widehat{x}$ by feeding the search region with adversarial perturbation $x+ \delta ^{adv}$ into the defense network parameterized by $\theta$.\\ $\widehat{x}=Def(x+\delta ^{adv},\theta)$
		
		\STATE After the second forward pass, recalculate the Dua-Loss with new $\widehat{x}$.\\ $L_{Dua}\left ( f\left ( z,\widehat{x} \right ),\left\{ y^{cls}, y^{reg} \right\} \right )$
		
		\STATE Compute the gradient of $L_{Dua}$ to defense network parameters $\theta$ and update $\theta$ with ADAM optimizer.

		\ENDFOR
		\ENDFOR
		
		\RETURN trained defense network parameters $\theta$

	\end{algorithmic}
\end{algorithm}

\section{Experiments}

\subsection{Experiment Settings}

\textbf{Implementation and Training Details.} The proposed AADN is implemented in PyTorch. The training phase was conducted on hardware consisting of an Intel i9-9900X CPU, 32GB RAM, and Nvidia RTX 3090 GPU. The network is pretrained on the ImageNet VID and COCO datasets. The training and testing process follows an end-to-end approach, where we fix the parameters of the tracker and update only the defense network using the ADAM optimizer. The ADAM's $\beta_{1}$ and $\beta_{2}$ are set to 0.5 and 0.999, respectively. The Dua-Loss weights $L_{cls}$ and $L_{reg}$ equally at 1. We choose SiamRPN++~\cite{siamrpnpp} tracker as our baseline tracker in training phase. The training process consists of 10 epochs, with a batch size of 16. The learning rate is set to 0.005. Adversarial training uses a perturbation budget of $\epsilon$ = 2.5/255. For the defense networks in the template branch and search branch, we employ UNet-128 and UNet-256 architectures, respectively. It is possible to configure the network to work simultaneously in both branches or selectively in one branch.  

\textbf{Test Dataset.}  We evaluate the proposed defense network on 3 widely used datasets: OTB100~\cite{otb2015}, VOT2018~\cite{vot2018} and LaSOT~\cite{lasot}. Specifically, OTB100 dataset consists of 100 challenging sequences captured from daily life scenes. Each sequence is labeled with 9 attributes that represent specific difficult scenes, such as illumination variation, scale variation, occlusion, deformation, motion blur, fast motion, and in-plane rotation. VOT2018 dataset is another challenging tracking benchmark that evaluates the accuracy and robustness of trackers. It consists of 60 videos and ranks the performance of trackers based on the expected average overlap (EAO) metric. LaSOT dataset is a recently released large-scale single object tracking benchmark. It comprises a total of 280 tracking sequences, featuring diverse challenging scenes and more than 70 classes of targets. It provides high-quality manual annotations for each frame.

\begin{table}[h!]
	\centering
	\caption{The accuracy performance of AADN against different adversarial attacks. \textit{Defense Pattern} denotes the deployment position of the defense network. \textit{Attack Result} denotes the original attack results without defense. \textit{Non-Adaptive Defense Result} represents the defense performance of the defense network against unadjusted adversarial attacks. \textit{Adaptive Defense Result} represents the defensive performance in the scenario of adaptive attacks. $\bigtriangleup_{non-adpt}$ and $\bigtriangleup_{adpt}$ indicate the changes in defense performance relative to the attack results. Enhancement results are highlighted in \color[HTML]{FE0000} \textbf{red} \color{black} and \color[HTML]{196f3d} \textbf{green} \color{black} based on the magnitude of improvement, respectively. }
	\label{table-attack}
	\scalebox{0.55}{
		\begin{tabular}{cccccccccc}
			\hline
			Dataset & Exp. No. & Attack Method & Defense Pattern & Metrics & Attack Result & \makecell{Non-Adaptive \\ Defense Result}  $\uparrow$ & $\bigtriangleup_{non-adpt}$(\%) & \multicolumn{1}{l}{\makecell{Adaptive \\ Defense Result} $\uparrow$} & $\bigtriangleup_{adpt}$(\%) \\ \hline

			\multirow{14}{*}{OTB100} & \multirow{2}{*}{1-1} & \multirow{6}{*}{CSA} & \multirow{2}{*}{\makecell{Template \\ Only}} & Success $\uparrow$ & 0.527 & \color[HTML]{FE0000}0.665 & \color[HTML]{FE0000}0.138(26.19\%) & \color[HTML]{196f3d}0.65 & \color[HTML]{196f3d}0.123(23.34\%) \\
			&  &  &  & Precision $\uparrow$ & 0.713 & \color[HTML]{FE0000}0.875 & \color[HTML]{FE0000}0.162(22.72\%) & \color[HTML]{196f3d}0.856 & \color[HTML]{196f3d}0.143(20.06\%) \\ \cline{2-2} \cline{4-10} 
			& \multirow{2}{*}{1-2} &  & \multirow{2}{*}{\makecell{Search \\ Only}} & Success $\uparrow$ & 0.349 & \color[HTML]{FE0000}0.561 & \color[HTML]{FE0000}0.212(60.74\%) & \color[HTML]{196f3d}0.461 & \color[HTML]{196f3d}0.112(32.09\%) \\
			&  &  &  & Precision $\uparrow$ & 0.491 & \color[HTML]{FE0000}0.766 & \color[HTML]{FE0000}0.275(56.01\%) & \color[HTML]{196f3d}0.636 & \color[HTML]{196f3d}0.145(29.53\%) \\ \cline{2-2} \cline{4-10} 
			& \multirow{2}{*}{1-3} &  & \multirow{2}{*}{Both} & Success $\uparrow$ & 0.324 & \color[HTML]{FE0000}0.559 & \color[HTML]{FE0000}0.235(72.53\%) & \color[HTML]{196f3d}0.403 & \color[HTML]{196f3d}0.079(24.38\%) \\
			&  &  &  & Precision $\uparrow$ & 0.471 & \color[HTML]{FE0000}0.777 & \color[HTML]{FE0000}0.396(64.97\%) & \color[HTML]{196f3d}0.573 & \color[HTML]{196f3d}0.102(21.66\%) \\ \cline{2-10} 
			& \multirow{2}{*}{1-4} & \multirow{2}{*}{IoU Attack} & \multirow{2}{*}{\makecell{Search \\ Only}} & Success $\uparrow$ & 0.499 & \color[HTML]{196f3d}0.618 & \color[HTML]{196f3d}0.119(23.85\%) & \color[HTML]{FE0000}0.627 & \color[HTML]{FE0000}0.128(25.65\%) \\ 
			&  &  &  & Precision $\uparrow$ & 0.644 & \color[HTML]{196f3d}0.824 & \color[HTML]{196f3d}0.18(27.95\%) & \color[HTML]{FE0000}0.826 & \color[HTML]{FE0000}0.182(28.26\%) \\ \cline{2-10}
			& \multirow{2}{*}{1-5} & \multirow{6}{*}{DFA} & \multirow{2}{*}{\makecell{Template\\Only}} & Success $\uparrow$ & 0.467 & \color[HTML]{FE0000}0.612 & \color[HTML]{FE0000}0.145(31.05\%) & \color[HTML]{196f3d}0.606 & \color[HTML]{196f3d}0.139(29.76\%) \\
			&  &  &  & Precision $\uparrow$ & 0.637 & \color[HTML]{FE0000}0.801 & \color[HTML]{FE0000}0.164(25.75\%) & \color[HTML]{196f3d}0.784 & \color[HTML]{196f3d}0.147(23.08\%) \\ \cline{2-2} \cline{4-10} 
			& \multirow{2}{*}{1-6} &  & \multirow{2}{*}{\makecell{Search\\Only}} & Success $\uparrow$ & 0.336 & \color[HTML]{FE0000}0.56 & \color[HTML]{FE0000} 0.224(66.67\%) & \color[HTML]{196f3d}0.486 & \color[HTML]{196f3d}0.15(44.64\%) \\
			&  &  &  & Precision $\uparrow$ & 0.517 & \color[HTML]{FE0000}0.794 & \color[HTML]{FE0000}0.277(53.58\%) & \color[HTML]{196f3d}0.663 & \color[HTML]{196f3d}0.146(28.24\%) \\ \cline{2-2} \cline{4-10} 
			& \multirow{2}{*}{1-7} &  & \multirow{2}{*}{Both} & Success $\uparrow$ & 0.355 & \color[HTML]{FE0000}0.555 &\color[HTML]{FE0000} 0.2(56.34\%) & \color[HTML]{196f3d}0.424 & \color[HTML]{196f3d}0.069(19.44\%) \\
			&  &  &  & Precision $\uparrow$ & 0.548 & \color[HTML]{FE0000}0.821 & \color[HTML]{FE0000}0.273(49.82\%) & \color[HTML]{196f3d}0.63 & \color[HTML]{196f3d}0.082(14.96\%) \\\hline

			\multirow{18}{*}{VOT2018} & \multirow{3}{*}{2-1} & \multirow{9}{*}{CSA} & \multirow{3}{*}{\makecell{Template \\ Only}} & EAO $\uparrow$ & 0.123 & \color[HTML]{FE0000}0.279 & \color[HTML]{FE0000}0.156(126.82\%) & \color[HTML]{196f3d}0.241 & \color[HTML]{196f3d}0.118(95.93\%) \\
			&  &  &  & Accuracy $\uparrow$ & 0.485 & \color[HTML]{196f3d}0.596 & \color[HTML]{196f3d}0.111(22.88\%) & \color[HTML]{FE0000}0.603 & \color[HTML]{FE0000}0.118(24.33\%) \\
			&  &  &  & Robustness $\downarrow$ & 2.145 & \color[HTML]{FE0000}0.44 & \color[HTML]{FE0000}1.705(79.48\%) & \color[HTML]{196f3d}0.524 & \color[HTML]{196f3d}1.621(75.57\%) \\ \cline{2-2} \cline{4-10} 
			& \multirow{3}{*}{2-2} &  & \multirow{3}{*}{\makecell{Search \\ Only}} & EAO $\uparrow$ & 0.069 & \color[HTML]{FE0000}0.154 & \color[HTML]{FE0000}0.085(123.18\%) & \color[HTML]{196f3d}0.104 & \color[HTML]{196f3d}0.035(50.72\%) \\
			&  &  &  & Accuracy $\uparrow$ & 0.541 & \color[HTML]{FE0000}0.557 & \color[HTML]{FE0000}0.016(2.95\%) & \color[HTML]{196f3d}0.549 & \color[HTML]{196f3d}0.008(1.48\%) \\
			&  &  &  & Robustness $\downarrow$ & 1.147 & \color[HTML]{FE0000}0.927 & \color[HTML]{FE0000}0.22(19.18\%) & \color[HTML]{196f3d}1.573 & \color[HTML]{196f3d}-0.426(-37.14\%) \\ \cline{2-2} \cline{4-10} 
			& \multirow{3}{*}{2-3} &  & \multirow{3}{*}{Both} & EAO $\uparrow$ & 0.073 & \color[HTML]{FE0000}0.14 & \color[HTML]{FE0000}0.067(91.78\%) & \color[HTML]{196f3d}0.109 & \color[HTML]{196f3d}0.036(49.32\%) \\
			&  &  &  & Accuracy $\uparrow$ & 0.467 & \color[HTML]{FE0000}0.546 & \color[HTML]{FE0000}0.079(16.91\%) & \color[HTML]{196f3d}0.488 & \color[HTML]{196f3d}0.021(4.50\%) \\
			&  &  &  & Robustness $\downarrow$ & 2.013 & \color[HTML]{FE0000}1.063 & \color[HTML]{FE0000}0.95(47.19\%) & \color[HTML]{196f3d}1.395 & \color[HTML]{196f3d}0.618(30.70\%) \\ \cline{2-10} 
			& \multirow{3}{*}{2-4} & \multirow{3}{*}{IoU Attack} & \multirow{3}{*}{\makecell{Search \\ Only}} & EAO $\uparrow$ & 0.129 & \color[HTML]{196f3d}0.265 & \color[HTML]{196f3d}0.136(105.42\%) & \color[HTML]{FE0000}0.334 & \color[HTML]{FE0000}0.205(158.91\%) \\
			&  &  &  & Accuracy $\uparrow$ & 0.568 & \color[HTML]{196f3d}0.588 & \color[HTML]{196f3d}0.02(3.52\%) & \color[HTML]{FE0000}0.647 & \color[HTML]{FE0000}0.079(13.91\%) \\
			&  &  &  & Robustness $\downarrow$ & 1.171 & \color[HTML]{196f3d}0.563 & \color[HTML]{196f3d}0.608(51.92\%) & \color[HTML]{FE0000}0.431 & \color[HTML]{FE0000}0.74(63.19\%) \\ \cline{2-10}
			& \multirow{3}{*}{2-5} & \multirow{9}{*}{DFA} & \multirow{3}{*}{\makecell{Template\\Only}} & EAO $\uparrow$ & 0.139 & \color[HTML]{FE0000}0.288 & \color[HTML]{FE0000}0.149(107.19\%) & \color[HTML]{196f3d}0.268 & \color[HTML]{196f3d}0.129(92.81\%) \\
			&  &  &  & Accuracy $\uparrow$ & 0.543 & \color[HTML]{FE0000}0.647 & \color[HTML]{FE0000}0.104(19.15\%) & \color[HTML]{196f3d}0.62 & \color[HTML]{196f3d}0.077(14.18\%) \\
			&  &  &  & Robustness $\downarrow$ & 1.105 & \color[HTML]{FE0000}0.371 & \color[HTML]{FE0000}0.734(66.43\%) & \color[HTML]{196f3d}0.414 & \color[HTML]{196f3d}0.691(62.53\%) \\ \cline{2-2} \cline{4-10} 
			& \multirow{3}{*}{2-6} &  & \multirow{3}{*}{\makecell{Search\\Only}} & EAO $\uparrow$ & 0.071 & \color[HTML]{FE0000}0.17 & \color[HTML]{FE0000}0.099(139.44\%) & \color[HTML]{196f3d}0.122 & \color[HTML]{196f3d}0.051(71.83\%) \\
			&  &  &  & Accuracy $\uparrow$ & 0.382 & \color[HTML]{FE0000}0.413 & \color[HTML]{FE0000}0.031(8.12\%) & \color[HTML]{196f3d}0.395 & \color[HTML]{196f3d}0.013(3.40\%) \\
			&  &  &  & Robustness $\downarrow$ & 2.056 & \color[HTML]{FE0000}1.003 & \color[HTML]{FE0000}1.053(51.22\%) & \color[HTML]{196f3d}1.161 & \color[HTML]{196f3d}0.895(43.53\%) \\ \cline{2-2} \cline{4-10} 
			& \multirow{3}{*}{2-7} &  & \multirow{3}{*}{Both} & EAO $\uparrow$ & 0.070 & \color[HTML]{FE0000}0.131 & \color[HTML]{FE0000}0.061(87.14\%) & \color[HTML]{196f3d}0.111 & \color[HTML]{196f3d}0.041(58.57\%) \\
			&  &  &  & Accuracy $\uparrow$ & 0.354 & \color[HTML]{FE0000}0.412 & \color[HTML]{FE0000}0.058(16.38\%) & \color[HTML]{196f3d}0.399 & \color[HTML]{196f3d}0.045(12.71\%) \\
			&  &  &  & Robustness $\downarrow$ & 1.892 & \color[HTML]{FE0000}0.985 & \color[HTML]{FE0000}0.907(47.94\%) & \color[HTML]{196f3d}1.017 & \color[HTML]{196f3d}0.875(46.25\%) \\ \hline

			\multirow{14}{*}{LaSOT} & \multirow{2}{*}{3-1} & \multirow{6}{*}{CSA} & \multirow{2}{*}{\makecell{Template \\ Only}} & Success $\uparrow$ & 0.393 & \color[HTML]{FE0000}0.489 & \color[HTML]{FE0000}0.096(24.42\%) & \color[HTML]{196f3d}0.471 & \color[HTML]{196f3d}0.078(19.85\%) \\
			&  &  &  & Norm. Precision $\uparrow$ & 0.448 & \color[HTML]{FE0000}0.569 & \color[HTML]{FE0000}0.121(27.00\%) & \color[HTML]{196f3d}0.548 & \color[HTML]{196f3d}0.1(22.32\%) \\ \cline{2-2} \cline{4-10} 
			& \multirow{2}{*}{3-2} &  & \multirow{2}{*}{\makecell{Search \\ Only}} & Success $\uparrow$ & 0.18 & \color[HTML]{FE0000}0.359 & \color[HTML]{FE0000}0.179(99.44\%) & \color[HTML]{196f3d}0.261 & \color[HTML]{196f3d}0.081(45.00\%) \\
			&  &  &  & Norm. Precision $\uparrow$ & 0.219 & \color[HTML]{FE0000}0.425 & \color[HTML]{FE0000}0.206(94.06\%) & \color[HTML]{196f3d}0.303 & \color[HTML]{196f3d}0.084(38.36\%) \\ \cline{2-2} \cline{4-10} 
			& \multirow{2}{*}{3-3} &  & \multirow{2}{*}{Both} & Success $\uparrow$ & 0.168 & \color[HTML]{FE0000}0.337 & \color[HTML]{FE0000}0.169(100.59\%) & \color[HTML]{196f3d}0.221 & \color[HTML]{196f3d}0.053(31.55\%) \\
			&  &  &  & Norm. Precision $\uparrow$ & 0.201 & \color[HTML]{FE0000}0.399 & \color[HTML]{FE0000}0.198(98.50\%) & \color[HTML]{196f3d}0.264 & \color[HTML]{196f3d}0.063(31.34\%) \\ \cline{2-10} 
			& \multirow{2}{*}{3-4} & \multirow{2}{*}{IoU Attack} & \multirow{2}{*}{\makecell{Search \\ Only}} & Success $\uparrow$ & 0.334 & \color[HTML]{196f3d}0.386 & \color[HTML]{196f3d}0.052(15.56\%) & \color[HTML]{FE0000}0.432 & \color[HTML]{FE0000}0.098(29.34\%) \\
			&  &  &  & Norm. Precision $\uparrow$ & 0.387 & \color[HTML]{196f3d}0.442 & \color[HTML]{196f3d}0.055(14.21\%) & \color[HTML]{FE0000}0.495 & \color[HTML]{FE0000}0.108(27.91\%) \\ \cline{2-10}
			& \multirow{2}{*}{3-5} & \multirow{6}{*}{DFA} & \multirow{2}{*}{\makecell{Template\\Only}} & Success $\uparrow$ & 0.369 & \color[HTML]{FE0000}0.477 & \color[HTML]{FE0000}0.108(29.27\%) & \color[HTML]{196f3d}0.444 & \color[HTML]{196f3d}0.075(20.33\%) \\
			&  &  &  & Norm. Precision $\uparrow$ & 0.316 & \color[HTML]{FE0000}0.426 & \color[HTML]{FE0000}0.11(34.81\%) & \color[HTML]{196f3d}0.387 & \color[HTML]{196f3d}0.071(22.47\%) \\ \cline{2-2} \cline{4-10} 
			& \multirow{2}{*}{3-6} &  & \multirow{2}{*}{\makecell{Search\\Only}} & Success $\uparrow$ & 0.161 & \color[HTML]{FE0000}0.338 & \color[HTML]{FE0000}0.177(109.94\%) & \color[HTML]{196f3d}0.256 & \color[HTML]{196f3d}0.095(59.01\%) \\
			&  &  &  & Norm. Precision $\uparrow$ & 0.124 & \color[HTML]{FE0000}0.257 & \color[HTML]{FE0000}0.133(107.26\%) & \color[HTML]{196f3d}0.197 & \color[HTML]{196f3d}0.073(58.87\%) \\ \cline{2-2} \cline{4-10} 
			& \multirow{2}{*}{3-7} &  & \multirow{2}{*}{Both} & Success $\uparrow$ & 0.17 & \color[HTML]{FE0000}0.357 & \color[HTML]{FE0000}0.187(110.00\%) & \color[HTML]{196f3d}0.234 & \color[HTML]{196f3d}0.064(37.65\%) \\
			&  &  &  & Norm. Precision $\uparrow$ & 0.166 & \color[HTML]{FE0000}0.344 & \color[HTML]{FE0000}0.178(107.23\%) & \color[HTML]{196f3d}0.218 & \color[HTML]{196f3d}0.052(31.33\%) \\ \hline
		\end{tabular}
	}

\end{table}

\subsection{Robustness on non-Adaptive Attacks}
\label{sec-attack}

To investigate the generalization of AADN against different adversarial attacks, we employ white-box attack method CSA (Exp.No.*-1, *-2, *-3), DFA (Exp.No.*-5, *-6, *-7), and black-box attack method IoU Attack (Exp.No.*-4) to generate adversarial perturbations. We then separately evaluate the defensive performance under different defense patterns (template branch only, search branch only, and both branch) using the OPE metrics. The results are presented in the column \textit{Non-Adaptive Defense Result} of \cref{table-attack}.

In terms of CSA and DFA, AADN demonstrates effective defense performance in all defense patterns, across 3 datasets. In (Exp.No.2-1), the defense network shows more significant gains in EAO and Robustness metrics, reaching 126.82\% and 79.48\%, respectively. This improves that AADN can effectively enhance the tracking accuracy and robustness of the tracker under adversarial attack conditions.

Since the IoU Attack is designed to target the search region, this part of the experiment is conducted on the search region branch. AADN demonstrates effective defense against IoU Attack on all datasets. (Exp.No.2-4) shows that the introduction of AADN can effectively mitigate the tracking failure and significantly improve the EAO metric.

\subsection{Robustness on Adaptive Attacks}

In recent years, increasing research~\cite{adaptive1,adaptive2,adaptive3} has indicated that static attacks (i.e., non-adaptive attacks in \cref{sec-attack}) against the original model are insufficient for evaluating the robustness of defense methods. In contrast, adaptive attacks aim to dynamically adjust the generation of adversarial samples based on the response from the defense mechanism. It is considered an important avenue for assessing the robustness of defense methods. Therefore, we conducted experiments on adaptive attacks.

Specifically, we combine AADN with the baseline tracker to create an integrated network. Then we retrain the CSA attack model(Exp.No.*-1, *-2, *-3) , DFA attack model(Exp.No.*-5, *-6, *-7) and utilize the loss information for IoU Attack(Exp.No.*-4) based on the integrated network to generate adversarial perturbation. The OPE evaluation in the adaptive attack scenario is presented in the column \textit{Adaptive Defense Result} of \cref{table-attack}.

Based on the results of (Exp.No.1-1, 1-2, 1-3) and (Exp.No.3-1, 3-2, 3-3), the proposed AADN maintains effective defense against adaptive CSA attack on the OTB100 and LaSOT datasets. However, compared to non-adaptive attack, it exhibits a decrease in defense performance. The reason behind this is that the attack models retrained specifically for the intergrated network can learn the defense mechanism and further enhance their targeted attack capabilities. This leads to a decline in defense performance.

In (Exp.No.2-2), AADN demonstrates a significant improvement in the EAO and Accuracy metrics on the search branch. However, there is a decrease in the Robustness metric. This indicates that in the scenarios of adaptive attack, the defense network can enhance the accuracy of the tracking results, the trade-off is that the tracker may exhibit target loss phenomena.

According to the results of (Exp.No.*-4), AADN reports impressive defense performance against adaptive IoU Attacks on all datasets. However, it can be observed that the results in the column \textit{Adaptive Defense Result} are superior to those in the  column \textit{Non-Adaptive Defense Result}. The reason for this is that the mechanism of IoU Attack involves iterative attacks on frames with an IoU score below a specific threshold, otherwise continuing to use adversarial perturbations from preceding frames. Due to the introduction of AADN, which improves the EAO and Accuracy metrics on the dataset, IoU Attack can only add adversarial perturbations from preceding frames where attacks have failed. As a result, the adaptive attacks yields better results compared to non-adaptive attacks.

\subsection{Generalization on Clean Samples}
\label{sec-clean}
In real-world scenarios, deployers of defense models often lack knowledge about whether the input images contain adversarial perturbations. Therefore, a defense network should not only filter out attack perturbations but also avoid significantly decreasing the accuracy of the tracker on clean samples. To address this, we input clean samples into both the AADN and the baseline tracker, corresponding results are presented in \cref{table-clean}.

\begin{table}[ht!]
	\centering
	\caption{The accuracy performance of AADN on clean samples. \textit{Original Result} represents the experimental results of the baseline tracker on clean samples without the addition of the defense network.}
	\label{table-clean}
	\scalebox{0.7}{
		\begin{tabular}{ccccccc}
			\hline
			Dataset & Exp. No. & Defense Pattern & Metrics & Original Result & Defense Result $\uparrow$ & $\bigtriangleup$(\%) \\ \hline
			\multirow{6}{*}{OTB100} & \multirow{2}{*}{1} & \multirow{2}{*}{Template Only} & Success $\uparrow$ & 0.695 & 0.672 & -0.024(-3.30\%) \\
			&  &  & Precision $\uparrow$ & 0.905 & 0.884 & -0.021(-2.32\%) \\ \cline{2-7} 
			& \multirow{2}{*}{2} & \multirow{2}{*}{Search Only} & Success $\uparrow$ & 0.695 & 0.64 & -0.055(-7.91\%) \\
			&  &  & Precision $\uparrow$ & 0.905 & 0.847 & -0.058(-6.40\%) \\ \cline{2-7} 
			& \multirow{2}{*}{3} & \multirow{2}{*}{Both} & Success $\uparrow$ & 0.695 & 0.644 & -0.051(-7.33\%) \\
			&  &  & Precision $\uparrow$ & 0.905 & 0.85 & -0.055(-6.07\%) \\ \hline
			\multirow{9}{*}{VOT2018} & \multirow{3}{*}{4} & \multirow{3}{*}{Template Only} & EAO $\uparrow$ & 0.352 & 0.343 & -0.009(-2.55\%) \\
			&  &  & Accuracy $\uparrow$ & 0.601 & 0.602 & 0.011(0.16\%) \\
			&  &  & Robustness $\downarrow$ & 0.29 & 0.3 & -0.01(-3.44\%) \\ \cline{2-7} 
			& \multirow{3}{*}{5} & \multirow{3}{*}{Search Only} & EAO $\uparrow$ & 0.352 & 0.245 & -0.107(-30.39\%) \\
			&  &  & Accuracy $\uparrow$ & 0.601 & 0.581 & -0.02(-3.32\%) \\
			&  &  & Robustness $\downarrow$ & 0.29 & 0.51 & -0.22(-75.86\%) \\ \cline{2-7} 
			& \multirow{3}{*}{6} & \multirow{3}{*}{Both} & EAO $\uparrow$ & 0.352 & 0.248 & -0.104(-29.54\%) \\
			&  &  & Accuracy $\uparrow$ & 0.601 & 0.592 & -0.009(-1.49\%) \\
			&  &  & Robustness $\downarrow$ & 0.29 & 0.51 & -0.22(-75.86\%) \\ \hline
			\multirow{6}{*}{LaSOT} & \multirow{2}{*}{7} & \multirow{2}{*}{Template Only} & Success $\uparrow$ & 0.496 & 0.501 & 0.005(1.00\%) \\
			&  &  & Norm. Precision $\uparrow$ & 0.575 & 0.575 & 0(0.00\%) \\ \cline{2-7} 
			& \multirow{2}{*}{8} & \multirow{2}{*}{Search Only} & Success $\uparrow$ & 0.496 & 0.432 & -0.064(-12.90\%) \\
			&  &  & Norm. Precision $\uparrow$ & 0.575 & 0.493 & -0.082(-14.26\%) \\ \cline{2-7} 
			& \multirow{2}{*}{9} & \multirow{2}{*}{Both} & Success $\uparrow$ & 0.496 & 0.455 & -0.041(-8.26\%) \\
			&  &  & Norm. Precision $\uparrow$ & 0.575 & 0.522 & -0.053(-9.21\%) \\ \hline
		\end{tabular}
	}

\end{table}

In the setting where the defense network operates solely on the template branch (Exp.No.1, 4, 7), the introduction of AADN does not significantly affect the tracking results. In fact, we even observe a slight improvement in tracking accuracy on the VOT2018 dataset (Exp.No.4) and the LaSOT dataset (Exp.No.7). In the setting where the defense network operates solely on the search branch (Exp.No.2, 5, 8), the defense network causes a limited loss in tracking accuracy and tracking robustness when facing clean samples. In the setting where the defense network operates on both branches simultaneously (Exp.No.3, 6, 9), the tracking results is better than only applying defense on the search branch. We posit that the transformed images through AADN lie outside the natural image manifold. If defense is applied only to one branch, it results in the template and search regions residing in different data manifolds, leading to performance degradation. However, deploying defense simultaneously ensures that both template and search images reside in the same data manifold, resulting in superior performance.

It is worth noting that we observed a significant drop in EAO on the VOT2018 dataset(Exp.No.5, 6). This can be attributed to an increase in tracking failure caused by applying defense on clean samples, resulting in a decrease in Robustness metric. Since the EAO metric evaluates the tracking accuracy (measured by the Accuracy metric) and tracking failure (measured by the Robustness metric), the substantial decrease in EAO is primarily driven by the Robustness. However, the AADN method performs well in terms of the Accuracy.

\subsection{Transferability on Different Visual Trackers}

In order to test the transferability of AADN, we transfer AADN directly to serveral trackers without retraining, including siamese trackers( SiamRPN\cite{siamrpn}, SiamMask\cite{siammask}), and anchor-free tracker OCEAN\cite{ocean}. We apply the IoU Attack method to introduce adversarial perturbations on these trackers seperately. Specifically, considering the structural characteristics of the IoU Attack method, we only perform experiments on its search region. The detailed experimental results are presented in \cref{table-tracker}.

The results demonstrate that even without retraining, AADN can still exhibit effective defense on the 3 trackers. Additionally, the results of (Exp.No.3) also showcase the effectiveness of AADN on anchor-free tracker. These findings highlight the excellent transferability of AADN and its ability to adapt to heterogeneous trackers.

\begin{table}[ht!]
	\centering
	\caption{The accuracy performance of AADN when transfered to different trackers without retraining.}
	\label{table-tracker}
	\scalebox{0.7}{
		\begin{tabular}{ccccccc}
			\hline
			Exp.No. & Baseline Tracker & Defense Pattern & Metrics & Attack Result & Defense Result $\uparrow$ & $\bigtriangleup$(\%) \\ \hline
			\multirow{2}{*}{1} & \multirow{2}{*}{SiamRPN} & \multirow{2}{*}{\makecell{Search\\Only}} & Success $\uparrow$ & 0.490 & 0.517 & 0.027(5.51\%) \\
			&  &  & Precision $\uparrow$ & 0.699 & 0.734 & 0.035(5.01\%) \\ \hline
			\multirow{2}{*}{2} & \multirow{2}{*}{SiamMask} & \multirow{2}{*}{\makecell{Search\\Only}} & Success $\uparrow$ & 0.533 & 0.556 & 0.023(4.32\%) \\
			&  &  & Precision $\uparrow$ & 0.73 & 0.754 & 0.024(3.29\%) \\ \hline
			\multirow{2}{*}{3} & \multirow{2}{*}{OCEAN} & \multirow{2}{*}{\makecell{Search\\Only}} & Success $\uparrow$ & 0.594 & 0.613 & 0.019(3.20\%) \\
			&  &  & Precision $\uparrow$ & 0.818 & 0.843 & 0.025(3.06\%) \\ \hline
		\end{tabular}
	}
	
\end{table}

\subsection{Speed of Different Defense Patterns}

While introducing a defense network to enhance the tracker's robustness against adversarial attack, it inevitably leads to a decrease in tracking efficiency. If the defense network has low processing efficiency, it can be disastrous for the real-time nature of the tracker. Therefore, we evaluate the time overhead of AADN in different defense patterns.

\begin{table}[h]
	\centering
	\caption{The speed performance of AADN.}
	\label{table-speed}
	\scalebox{0.8}{
		\begin{tabular}{ccccc}
			\hline
			Exp. No. & Defense Pattern & Speed/FPS & Inference Time/ms & $\bigtriangleup$Time \\ \hline
			1 & Original SiamRPN++ & 107 & 9.35 & - \\
			2 & Template Only & 104 & 9.62 & -0.27 \\
			3 & Search Only & 70 & 14.29 & -4.94 \\
			4 & Both & 68 & 14.71 & -5.36 \\ \hline
		\end{tabular}
	}

\end{table}

As shown in the \cref{table-speed}, in (Exp.No.1), we measure the average tracking speed of SiamRPN++ to be 9.35 ms/frame. Combining the experimental results from (Exp.No.2), we obtain a performance overhead of 0.27 ms/frame for the template branch. Similarly, the performance loss for the search branch is measured to be 4.94 ms/frame. When deploying in both branches simultaneously, the defense network results in a performance decrease of 5.36 ms/frame. Therefore, when working in conjunction with other real-time trackers, the introduction of AADN does not significantly reduce the tracking speed while significantly improving the accuracy and robustness of the visual tracker against adversarial attack.

\subsection{Ablation Study on Dua-Loss}

In \cref{sec-dualoss}, we introduce the Dua-Loss, which consists of classification loss $L_{cls}$ and regression loss $L_{reg}$. To investigate the individual contributions of $L_{cls}$ and $L_{reg}$ to the network's defense performance, we separately train the defense network using these two loss functions and evaluated their corresponding defense effectiveness against the CSA attack on OTB100 dataset. The detailed experimental results are presented in \cref{table-loss}.

\begin{table}[h!]
	\centering
	\caption{The accuracy performance of AADN with different losses during adversarial training. The best results are highlighted in \color[HTML]{FE0000} \textbf{red}.}
	\label{table-loss}
	\scalebox{0.7}{
		\begin{tabular}{ccccccc}
			\hline
			Exp.No. & Defense Pattern & Loss Type & Metrics & Attack Result & Defense Result $\uparrow$ & $\bigtriangleup$(\%) \\ \hline
			&  &  & Success $\uparrow$ & 0.527 & {\color[HTML]{FE0000} 0.665} & {\color[HTML]{FE0000} 0.138(26.19\%)} \\
			\multirow{-2}{*}{1-1} &  & \multirow{-2}{*}{Dua-Loss} & Precision $\uparrow$ & 0.713 & {\color[HTML]{FE0000} 0.875} & {\color[HTML]{FE0000} 0.162(22.72\%)} \\ \cline{1-1} \cline{3-7} 
			&  &  & Success $\uparrow$ & 0.527 & 0.657 & 0.13(24.67\%) \\
			\multirow{-2}{*}{1-2} &  & \multirow{-2}{*}{$L_{cls}$} & Precision $\uparrow$ & 0.713 & 0.861 & 0.148(20.76\%) \\ \cline{1-1} \cline{3-7} 
			&  &  & Success $\uparrow$ & 0.527 & 0.639 & 0.112(21.25\%) \\
			\multirow{-2}{*}{1-3} & \multirow{-6}{*}{\makecell{Template \\ Only}} & \multirow{-2}{*}{$L_{reg}$} & Precision $\uparrow$ & 0.713 & 0.842 & 0.129(18.09\%) \\ \hline
			&  &  & Success $\uparrow$ & 0.349 & {\color[HTML]{FE0000} 0.561} & {\color[HTML]{FE0000} 0.212(60.74\%)} \\
			\multirow{-2}{*}{2-1} &  & \multirow{-2}{*}{Dua-Loss} & Precision $\uparrow$ & 0.491 & {\color[HTML]{FE0000} 0.766} & {\color[HTML]{FE0000} 0.275(56.01\%)} \\ \cline{1-1} \cline{3-7} 
			&  &  & Success $\uparrow$ & 0.349 & 0.554 & 0.205(58.74\%) \\
			\multirow{-2}{*}{2-2} &  & \multirow{-2}{*}{$L_{cls}$} & Precision $\uparrow$ & 0.491 & 0.747 & 0.256(52.14\%) \\ \cline{1-1} \cline{3-7} 
			&  &  & Success $\uparrow$ & 0.349 & 0.539 & 0.19(54.44\%) \\
			\multirow{-2}{*}{2-3} & \multirow{-6}{*}{\makecell{Search \\ Only}} & \multirow{-2}{*}{$L_{reg}$} & Precision $\uparrow$ & 0.491 & 0.743 & 0.252(51.32\%) \\ \hline
			&  &  & Success $\uparrow$ & 0.324 & {\color[HTML]{FE0000} 0.559} & {\color[HTML]{FE0000} 0.235(72.53\%)} \\
			\multirow{-2}{*}{3-1} &  & \multirow{-2}{*}{Dua-Loss} & Precision $\uparrow$ & 0.471 & {\color[HTML]{FE0000} 0.777} & {\color[HTML]{FE0000} 0.306(64.97\%)} \\ \cline{1-1} \cline{3-7} 
			&  &  & Success $\uparrow$ & 0.324 & 0.55 & 0.226(69.75\%) \\
			\multirow{-2}{*}{3-2} &  & \multirow{-2}{*}{$L_{cls}$} & Precision $\uparrow$ & 0.471 & 0.762 & 0.291(61.78\%) \\ \cline{1-1} \cline{3-7} 
			&  &  & Success $\uparrow$ & 0.324 & 0.545 & 0.221(68.21\%) \\
			\multirow{-2}{*}{3-3} & \multirow{-6}{*}{Both} & \multirow{-2}{*}{$L_{reg}$} & Precision $\uparrow$ & 0.471 & 0.757 & 0.286(60.72\%) \\ \hline
		\end{tabular}
	}
	
\end{table}

In all 3 defense patterns, using Dua-Loss for adversarial training (Exp.No. *-1) achieves the best defense performance. Subsequently, $L_{cls}$ and $L_{reg}$ result in a gradual decrease in defense effectiveness. One possible reason for this is that the tracking network's prediction heads use the classification branch for coarse localization of the target and the regression branch for fine localization. The adversarial samples generated by Dua-Loss can simultaneously attack both the classification and regression branches, making them more aggressive. Therefore, adversarial training guided by Dua-Loss achieves the best defense results. Furthermore, the results of (Exp.No.*-2) and (Exp.No.*-3) confirm that the attack effectiveness of adversarial samples generated by $L_{cls}$ and $L_{reg}$, respectively, decreases compared to Dua-Loss. As a result, the defense performance also decreases accordingly.

\section{Conclusion}

In this paper, we propose AADN, a auxiliary network for visual tracker, which enables adversarial defense for both the tracking template and search region. Extensive experimental results demonstrate that AADN exhibits good generalization against adversarial attack in both non-adaptive and adaptive attack scenarios. It can also transfer its defense effectiveness to other trackers. Furthermore, AADN also demonstrates excellent efficiency, allowing it to be used in conjunction with other high-speed trackers without introducing excessive computational overhead.

\section*{Acknowledgements}
This work was supported in part by the National Natural Science Foundation of China (62272089) and the Open Project of the Intelligent Terminal Key Laboratory of Sichuan Province (SCITLAB-30003).

%
%
\bibliographystyle{splncs04}
\bibliography{main}

\begin{thebibliography}{10}
\providecommand{\url}[1]{\texttt{#1}}
\providecommand{\urlprefix}{URL }
\providecommand{\doi}[1]{https://doi.org/#1}

\bibitem{adaptive3}
Athalye, A., Carlini, N., Wagner, D.: Obfuscated gradients give a false sense
  of security: Circumventing defenses to adversarial examples. In:
  International conference on machine learning. pp. 274--283. PMLR (2018)

\bibitem{atreview}
Bai, T., Luo, J., Zhao, J., Wen, B., Wang, Q.: Recent advances in adversarial
  training for adversarial robustness. arXiv preprint arXiv:2102.01356  (2021)

\bibitem{adaptive2}
Carlini, N., Wagner, D.: Adversarial examples are not easily detected:
  Bypassing ten detection methods. In: Proceedings of the 10th ACM workshop on
  artificial intelligence and security. pp. 3--14 (2017)

\bibitem{cw}
Carlini, N., Wagner, D.: Towards evaluating the robustness of neural networks.
  In: 2017 ieee symposium on security and privacy (sp). pp. 39--57. Ieee (2017)

\bibitem{oneshot}
Chen, X., Yan, X., Zheng, F., Jiang, Y., Xia, S.T., Zhao, Y., Ji, R.: One-shot
  adversarial attacks on visual tracking with dual attention. In: Proceedings
  of the IEEE/CVF conference on computer vision and pattern recognition. pp.
  10176--10185 (2020)

\bibitem{lasot}
Fan, H., Lin, L., Yang, F., Chu, P., Deng, G., Yu, S., Bai, H., Xu, Y., Liao,
  C., Ling, H.: Lasot: A high-quality benchmark for large-scale single object
  tracking. In: Proceedings of the IEEE/CVF Conference on Computer Vision and
  Pattern Recognition. pp. 5374--5383 (2019)

\bibitem{gramask}
Folz, J., Palacio, S., Hees, J., Dengel, A.: Adversarial defense based on
  structure-to-signal autoencoders. In: 2020 IEEE Winter Conference on
  Applications of Computer Vision (WACV). pp. 3568--3577. IEEE (2020)

\bibitem{fastrcnn}
Girshick, R.: Fast r-cnn. In: Proceedings of the IEEE international conference
  on computer vision. pp. 1440--1448 (2015)

\bibitem{fgsm}
Goodfellow, I.J., Shlens, J., Szegedy, C.: Explaining and harnessing
  adversarial examples. In: Bengio, Y., LeCun, Y. (eds.) 3rd International
  Conference on Learning Representations, {ICLR} 2015, San Diego, CA, USA, May
  7-9, 2015, Conference Track Proceedings (2015),
  \url{http://arxiv.org/abs/1412.6572}

\bibitem{spark}
Guo, Q., Xie, X., Juefei-Xu, F., Ma, L., Li, Z., Xue, W., Feng, W., Liu, Y.:
  Spark: Spatial-aware online incremental attack against visual tracking. In:
  Vedaldi, A., Bischof, H., Brox, T., Frahm, J.M. (eds.) Computer Vision --
  ECCV 2020. pp. 202--219. Springer International Publishing, Cham (2020)

\bibitem{iouattack}
Jia, S., Song, Y., Ma, C., Yang, X.: Iou attack: Towards temporally coherent
  black-box adversarial attack for visual object tracking. In: Proceedings of
  the IEEE/CVF Conference on Computer Vision and Pattern Recognition (CVPR).
  pp. 6709--6718 (June 2021)

\bibitem{vot2018}
Kristan, M., Leonardis, A., Matas, J., Felsberg, M., Pflugfelder, R.,
  ˇCehovin~Zajc, L., Vojir, T., Bhat, G., Lukezic, A., Eldesokey, A., et~al.:
  The sixth visual object tracking vot2018 challenge results. In: Proceedings
  of the European conference on computer vision (ECCV) workshops. pp.~0--0
  (2018)

\bibitem{bim}
Kurakin, A., Goodfellow, I.J., Bengio, S.: Adversarial examples in the physical
  world. In: Artificial intelligence safety and security, pp. 99--112. Chapman
  and Hall/CRC (2018)

\bibitem{siamrpnpp}
Li, B., Wu, W., Wang, Q., Zhang, F., Xing, J., Yan, J.: Siamrpn++: Evolution of
  siamese visual tracking with very deep networks. In: Proceedings of the
  IEEE/CVF conference on computer vision and pattern recognition. pp.
  4282--4291 (2019)

\bibitem{siamrpn}
Li, B., Yan, J., Wu, W., Zhu, Z., Hu, X.: High performance visual tracking with
  siamese region proposal network. In: Proceedings of the IEEE conference on
  computer vision and pattern recognition. pp. 8971--8980 (2018)

\bibitem{autodrive1}
Li, P., Jin, J.: Time3d: End-to-end joint monocular 3d object detection and
  tracking for autonomous driving. In: Proceedings of the IEEE/CVF Conference
  on Computer Vision and Pattern Recognition (CVPR). pp. 3885--3894 (June 2022)

\bibitem{uta}
Li, Z., Shi, Y., Gao, J., Wang, S., Li, B., Liang, P., Hu, W.: A simple and
  strong baseline for universal targeted attacks on siamese visual tracking.
  IEEE Transactions on Circuits and Systems for Video Technology
  \textbf{32}(6),  3880--3894 (2022). \doi{10.1109/TCSVT.2021.3120479}

\bibitem{fan}
Liang, S., Wei, X., Yao, S., Cao, X.: Efficient adversarial attacks for visual
  object tracking. In: Computer Vision--ECCV 2020: 16th European Conference,
  Glasgow, UK, August 23--28, 2020, Proceedings, Part XXVI 16. pp. 34--50.
  Springer (2020)

\bibitem{eusa}
Liu, S., Chen, Z., Li, W., Zhu, J., Wang, J., Zhang, W., Gan, Z.: Efficient
  universal shuffle attack for visual object tracking. In: ICASSP 2022 - 2022
  IEEE International Conference on Acoustics, Speech and Signal Processing
  (ICASSP). pp. 2739--2743 (2022). \doi{10.1109/ICASSP43922.2022.9747773}

\bibitem{autodrive2}
Luo, C., Yang, X., Yuille, A.: Exploring simple 3d multi-object tracking for
  autonomous driving. In: Proceedings of the IEEE/CVF International Conference
  on Computer Vision (ICCV). pp. 10488--10497 (October 2021)

\bibitem{pgdat}
Madry, A., Makelov, A., Schmidt, L., Tsipras, D., Vladu, A.: Towards deep
  learning models resistant to adversarial attacks. arXiv preprint
  arXiv:1706.06083  (2017)

\bibitem{defdistillation}
Papernot, N., McDaniel, P., Wu, X., Jha, S., Swami, A.: Distillation as a
  defense to adversarial perturbations against deep neural networks. In: 2016
  IEEE symposium on security and privacy (SP). pp. 582--597. IEEE (2016)

\bibitem{unet}
Ronneberger, O., Fischer, P., Brox, T.: U-net: Convolutional networks for
  biomedical image segmentation. In: Medical Image Computing and
  Computer-Assisted Intervention--MICCAI 2015: 18th International Conference,
  Munich, Germany, October 5-9, 2015, Proceedings, Part III 18. pp. 234--241.
  Springer (2015)

\bibitem{robot1}
Sandoval, L.A.C.: Low cost object tracking by computer vision using 8 bits
  communication with a viper robot. In: 2023 8th International Conference on
  Control and Robotics Engineering (ICCRE). pp. 232--237 (2023).
  \doi{10.1109/ICCRE57112.2023.10155618}

\bibitem{freeat}
Shafahi, A., Najibi, M., Ghiasi, M.A., Xu, Z., Dickerson, J., Studer, C.,
  Davis, L.S., Taylor, G., Goldstein, T.: Adversarial training for free! In:
  Wallach, H., Larochelle, H., Beygelzimer, A., d\textquotesingle
  Alch\'{e}-Buc, F., Fox, E., Garnett, R. (eds.) Advances in Neural Information
  Processing Systems. vol.~32. Curran Associates, Inc. (2019),
  \url{https://proceedings.neurips.cc/paper_files/paper/2019/file/7503cfacd12053d309b6bed5c89de212-Paper.pdf}

\bibitem{dfa}
Suttapak, W., Zhang, J., Zhang, L.: Diminishing-feature attack: The adversarial
  infiltration on visual tracking. Neurocomputing  \textbf{509},  21--33 (2022)

\bibitem{szegedy2014}
Szegedy, C., Zaremba, W., Sutskever, I., Bruna, J., Erhan, D., Goodfellow,
  I.J., Fergus, R.: Intriguing properties of neural networks. In: Bengio, Y.,
  LeCun, Y. (eds.) 2nd International Conference on Learning Representations,
  {ICLR} 2014, Banff, AB, Canada, April 14-16, 2014, Conference Track
  Proceedings (2014), \url{http://arxiv.org/abs/1312.6199}

\bibitem{adaptive1}
Tramer, F., Carlini, N., Brendel, W., Madry, A.: On adaptive attacks to
  adversarial example defenses. Advances in neural information processing
  systems  \textbf{33},  1633--1645 (2020)

\bibitem{siammask}
Wang, Q., Zhang, L., Bertinetto, L., Hu, W., Torr, P.H.: Fast online object
  tracking and segmentation: A unifying approach. In: Proceedings of the
  IEEE/CVF conference on Computer Vision and Pattern Recognition. pp.
  1328--1338 (2019)

\bibitem{robot2}
Wilson, J., Lin, M.C.: Avot: Audio-visual object tracking of multiple objects
  for robotics. In: 2020 IEEE International Conference on Robotics and
  Automation (ICRA). pp. 10045--10051 (2020).
  \doi{10.1109/ICRA40945.2020.9197528}

\bibitem{fastat}
Wong, E., Rice, L., Kolter, J.Z.: Fast is better than free: Revisiting
  adversarial training. In: 8th International Conference on Learning
  Representations, {ICLR} 2020, Addis Ababa, Ethiopia, April 26-30, 2020.
  OpenReview.net (2020), \url{https://openreview.net/forum?id=BJx040EFvH}

\bibitem{otb2015}
{Wu}, Y., {Lim}, J., {Yang}, M.: Object tracking benchmark. IEEE Transactions
  on Pattern Analysis and Machine Intelligence  \textbf{37}(9),  1834--1848
  (2015). \doi{10.1109/TPAMI.2014.2388226}

\bibitem{csa}
Yan, B., Wang, D., Lu, H., Yang, X.: Cooling-shrinking attack: Blinding the
  tracker with imperceptible noises. In: Proceedings of the IEEE/CVF Conference
  on Computer Vision and Pattern Recognition (CVPR) (June 2020)

\bibitem{ocean}
Zhang, Z., Peng, H., Fu, J., Li, B., Hu, W.: Ocean: Object-aware anchor-free
  tracking. In: Computer Vision--ECCV 2020: 16th European Conference, Glasgow,
  UK, August 23--28, 2020, Proceedings, Part XXI 16. pp. 771--787. Springer
  (2020)

\end{thebibliography}
\end{document}